\documentclass[letterpaper, 10 pt, conference]{ieeeconf}

\IEEEoverridecommandlockouts

\usepackage{amssymb}
\usepackage{makeidx}         
\usepackage{graphicx}        
\usepackage{float}
\usepackage[outercaption]{sidecap}
\usepackage{subfig}

\usepackage{epstopdf}
\usepackage{url}
\usepackage{wrapfig}
\usepackage{lipsum}

\usepackage{amssymb}
\usepackage{multicol}        
\usepackage{multirow}
\usepackage{hhline}
\usepackage{xcolor,colortbl}
\usepackage{float}
\usepackage{color,soul}
\usepackage[T1]{fontenc}
\usepackage{xcolor}
\usepackage{newverbs}

\usepackage{calc}
\usepackage{indentfirst}
\usepackage{fancyhdr}
\usepackage{graphicx,epstopdf}
\usepackage{lastpage}
\usepackage{ifthen}
\usepackage{lineno}
\usepackage{float}
\usepackage[tbtags]{amsmath}
\usepackage{setspace}
\usepackage{harpoon}
\usepackage{mathpazo}
\usepackage{booktabs} 
\usepackage{etoolbox} 
\usepackage{endnotes}
\usepackage[framemethod=tikz]{mdframed}
\usepackage{tikz}
\usepackage{algorithm2e}
\usepackage{longtable}
\usepackage{comment}
\usepackage[utf8]{inputenc}
\usepackage{array}
\usepackage{mathtools}
\newsavebox{\measurebox}
\usepackage{mwe}    
\usepackage{subfig}
\newcolumntype{P}[1]{>{\centering\arraybackslash}p{#1}}
\usetikzlibrary{
  matrix,
  positioning,
  decorations,
  decorations.pathreplacing
}
\newcommand {\Ref}[1] {~\cite{#1}}

\newcommand {\fig}[1] {Figure \ref{#1}}   
 
\newcommand{\tablefont}[1]{\textlf}
\DeclarePairedDelimiter{\norm}{\lVert}{\rVert}
\newcommand{\vectorproj}[2][]{\textit{proj}_{\vect{#1}}\vect{#2}}
\newcommand{\vect}{\mathbf}
\makeatletter
\newcommand*\bigcdot{\mathpalette\bigcdot@{.5}}
\newcommand*\bigcdot@[2]{\mathbin{\vcenter{\hbox{\scalebox{#2}{$\m@th#1\bullet$}}}}}
\makeatother

\DeclareMathOperator{\atantwo}{atan2}

\title{Real-time Joint Motion Analysis and Instrument Tracking \\for Robot-Assisted Orthopaedic Surgery}
\author{Mario Strydom$^{1,2}$, Artur Banach$^{1,2}$, Liao Wu$^{2,4}$, Ross Crawford$^{1,2,3}$,\\ Jonathan Roberts$^{1,2}$ and Anjali Jaiprakash$^{1,2}$
\thanks{*This work was supported by the Advanced Queensland Scheme and the Queensland University of Technology Research Training Program}
\thanks{$^{1}$Science and Engineering Faculty,Queensland University of Technology, Brisbane, Australia}
\thanks{$^{2}$Australian Centre for Robotic Vision}%
\thanks{$^{3}$The Prince Charles Hospital, Brisbane, Australia}
\thanks{$^{4}$School of Mechanical and Manufacturing Engineering, University of New South Wales, Sydney, Australia}
}
\begin{document}
\maketitle
\thispagestyle{empty}
\pagestyle{empty}
\begin{abstract}
Robotic-assisted orthopaedic surgeries demand accurate, automated leg manipulation for improved spatial accuracy to reduce iatrogenic damage. In this study, we propose novel rigid body designs and an optical tracking volume setup for tracking of the femur, tibia and surgical instruments. Anatomical points inside the leg are measured using Computed Tomography with an accuracy of 0.3mm. Combined with kinematic modelling, we can express these points relative to any frame and across joints to sub-millimetre accuracy. It enables the setup of vectors on the mechanical axes of the femur and tibia for kinematic analysis. Cadaveric experiments are used to verify the tracking of internal anatomies and joint motion analysis. The proposed integrated solution is a first step in the automation of leg manipulation and can be used as a ground-truth for future robot-assisted orthopaedic research.
\end{abstract}
\section{Introduction}
Over the last two decades, robot-assisted procedures have become a worldwide standard in surgical theatres. Robotic platforms have been deployed in a range of orthopaedic procedures. However, for minimally invasive surgeries (MIS) such as knee arthroscopy, little automation has been introduced, with a high level of iatrogenic damage\Ref{strydom2016towards}. 
\par
\begin{figure}[t]
\centering
    \subfloat[Single camera system showing surgeon need to move out of the optical path, adding time to the surgery.]{\label{HolySurgery}{\includegraphics[width=0.55\linewidth]{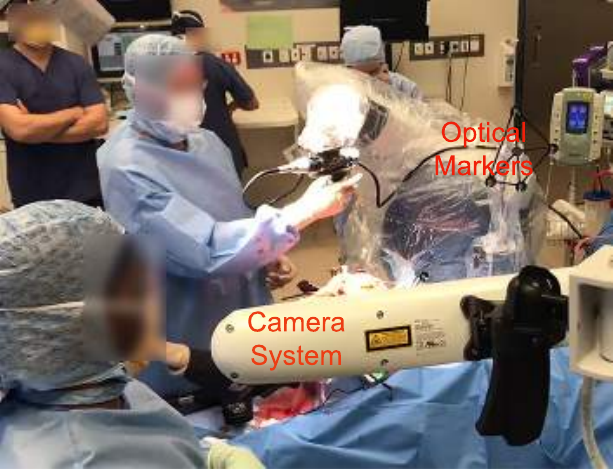}}}
    ~
    \subfloat[Broken OptiTrack marker on Arthroscope (right side).]{\label{BrokenMarkers}{\includegraphics[width=0.36\linewidth]{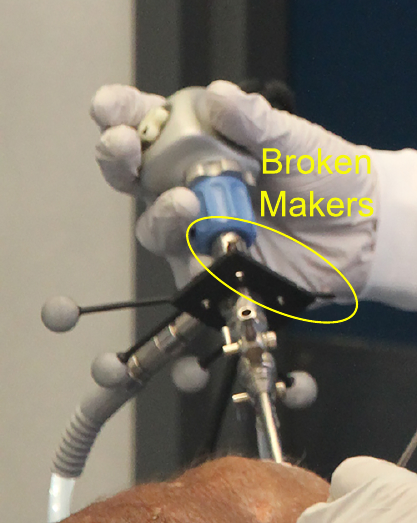}}}\\
         \caption{Surgical Delays and Optical Tracking Issues}
    \label{MarkerIssues}      
\end{figure}

Robotic-assisted surgery and automated leg manipulation demands real-time knowledge of the leg's pose. Optical tracking is today extensively used in operating theatres, however with the latest systems such as the Mako Rio \cite{Allen2019}, the optical system limits the surgeon's movement as shown in \fig{HolySurgery} and from observations increases the surgical time on average by 17\% to ensure optical tracking.
To mount the optical markers, surgical pins are drilled into the leg for selected procedures and rigid bodies are mounted on these pins. For minimally invasive surgeries, no tracking of the leg is done, and as a result no leg pose information is available. Existing rigid bodies are large as seen in \fig{HolySurgery} and interferes with the surgery. 
The standard rigid bodies from OptiTrack were initially tested and failed physically within a few minutes during the first cadaver arthroscopy as shown in \fig{BrokenMarkers}.
Modern systems use a single camera system, mounted on a tripod next to the patient (\fig{HolySurgery}), resulting in significant interference from surgical staff and equipment. During an actual arthroscopy, a slight movement was observed between the large RBs and the patient's leg and for measuring the small spaces inside the knee joint, the RBs need to be rigid relative to the leg for sub-millimetre accuracy.
For minimal invasive surgery such as an arthroscopy, surgeons physically position and restrict parts of the patient's leg in the coronal, sagittal or transverse planes to allow surgical equipment to reach specific positions inside the body (\fig{MerfSurgery}). For knee replacements without robotic support, they manually align the femur and tibia with varying accuracy levels that depends on experience. To control the nine Degrees of Freedom (DoF) \cite[p.~245]{Nigg} of the combined hip \cite{Reinbolt2005DeterminationOptimization} and knee motion \cite{Blankevoort1988TheMotion} it is necessary to estimate the poses of these joints in real-time accurately.
\par
Rigid bodies (RB) with markers (optical tracking balls) in specific positions are mounted on the patient's leg, as shown in \fig{LegRobot} and tracked using an optical tracking (OptiTrack) system. The position of the markers relative to a point inside the leg is measured using Computed Tomography (CT) scans of the leg. The leg pose can be calculated using the tracking data and local measurements.
Three cadaver experiments are used to refine the camera volume around the patient, and new rigid body designs are tested to support continuous tracking of optical markers irrespective of surgical staff movement.
For automated leg manipulation (and for future robotic surgery), robotic principals need to be applied to the human body. Mathematical solutions are developed in conjunction with the optical tracking system to track any point inside the leg. It enables continuous tracking of any position of the leg or joints to provide in real-time the pose of the leg with millimetre accuracy. A range of applications are enabled for robotic surgery, and the calculations of leg pose angles are developed to illustrate the impact of the integrated solution. A prototype surgical robot was developed and patented for automating leg manipulation, and used in this study as shown in \fig{LegRobot} to test the RBs.
\par
The contributions of this study focus on automating leg manipulation for future robotic leg surgery. It includes the optimisation of the optical tracking volume, development of novel rigid body (RB) designs and a mathematical framework for optical tracking and positional analysis of femur, tibia and surgical instruments. CT scanned cadaver images are used to analyse anatomical positions inside the leg, which enable us to calculate the leg pose and joint angles. A foot interface attaching the robot to the leg was developed to lock the ankle, manipulate the leg and remove surgical pins from the tibia.
\subsection{Assumptions}
A commercial optical tracking system (OptiTrack) with sub-millimetre accuracy was used during this study to test markers, rigid bodies and the kinematic model \cite{OptiTrack}. The system is not theatre ready, but provides a benchmark in a cadaveric experiment.
\subsection{Related Work}
From the hip to the heel, the leg has twenty-one DoF - three in the hip ball joint, six in the knee and effectively twelve DoF across the ankle joint complex (AJC) \cite{deAsla2006SixTechnique}. The hip is a ball and socket joint \cite{Reinbolt2005DeterminationOptimization}. Apkarian {\textit{et al.}} noticed that the hip joint leads through the femur head onto the femur with an offset (femoral neck), changing the rotational properties to extend the motion capabilities of the human's hip kinematic range \cite{Apkarian1989ALimb}. The knee joint is the largest joint in the human body with six DoF, three rotations and three translations \cite{Scuderi2010a}. For surgical applications, most of these variables are manipulated to gain access to the inner knee \cite{Scuderi2010a, Li2009}.
\par
Lu \textit{et al.} proposed an optimisation method for optical tracking of markers (optical reflector) mounted on the skin to determine the positions and orientations of multi-link musculoskeletal models from marker coordinates \cite{lu1999bone}. Their optimised model compensates for the movement of the markers on the skin, however not sufficient for use in surgeries where millimetre level accuracy is required.
Maletsky \textit{et al.} shows the accuracy of motion between two rigid bodies for biological experiments (translation and rotation); using an OptiTrack optical motion capture system; to be smaller than 0.1 (mm or deg) for a volume size of less than 4m \cite{Maletsky2007}. However, they don't address specific rigid body designs, and marker and volume configurations required for real-time tracking during surgery. More cameras will be required to ensure continuous coverage during surgery in a real surgical volume. Nagymate \textit{et al.} introduced a novel micro-triangulation-based validation and calibration method for motion capture systems, supporting the volume setup and calibration used in this study \cite{Nagymate2018}.
Yang \textit{et al.} proposed a design of an Optical system to track surgical tools, accurate to 0.009mm, however as with the Mako Rio system, their solution use only two cameras which is not suitable for real-time tracking with continuous leg motion \Ref{Yang2013}. 
\begin{figure}[t]
\centering
	\vspace{2 mm}
    \subfloat[A surgeon moving the leg with his body while performing a knee arthroscopy using optical tracking of the arthroscope.]{\label{MerfSurgery}{\includegraphics[width=0.56\linewidth]{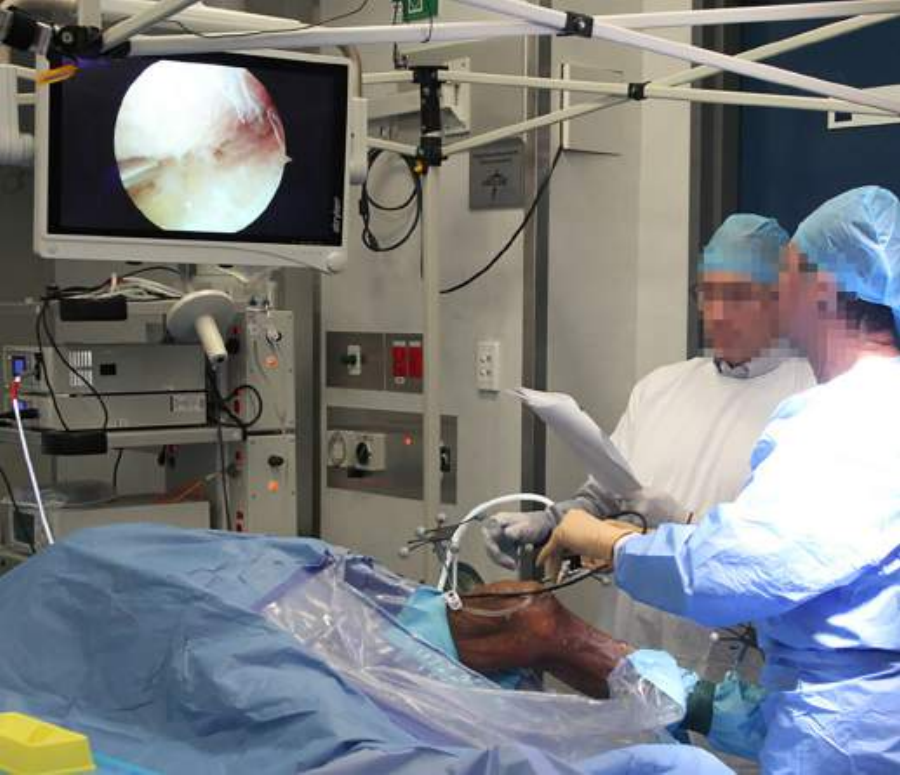}}}
    ~
    \subfloat[Prototype leg manipulation robot.]{\label{LegRobot}{\includegraphics[width=0.325\linewidth]{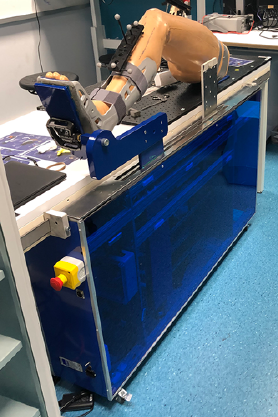}}}\\
         \caption{Towards Robotic Leg Manipulation}
    \label{CadExp}      
\end{figure} 
Computed Tomography (CT) is pre-operatively used in many orthopaedic surgeries (such as Mako) to measure or register points inside the leg. Kim \textit{et al.} showed that with CT scans, distances inside the body can be measured with an accuracy of 0.3mm \cite{kim2012accuracy}. In this study CT will be used to measure vectors from a marker on a RB to a point inside the leg.
\par
 Charlton \textit{\textit{et al.}} investigated the repeatability of an optimised lower body model as a measure of RB rigidity \cite{Charlton2004}. They, however, use a 3 DoF model for the knee and ankle. For robotic knee and hip surgery, all six degrees in the knee needs to be taken into consideration, where the translation is a few millimetres, especially in the posterior/anterior direction. For robotic leg manipulation the ankle is locked and the foot used to manipulate the leg. In their study, they constrained the coordinate systems to ensure minimal singularities, which is not done in this study. They further use skin markers, resulting in a low marker accuracy during motion, which is not feasible for robotic knee or hip surgery \cite{Charlton2004}.
 Finding the centre of the hip joint ball centre (HJC) is essential to measure the leg parameters accurately. Kainz \textit{\textit{et al.}}  provided a review of 34 articles on hip joint centre (HJC) estimation in human motion \cite{Kainz2015}. However, due to surgical accuracy required to track both the leg parameters and surgical instrument position, we show that a personalised and more accurate approach is to use customised optical rigid bodies, CT scan data and optical reference markers in combination with a kinematic transformation model to tune key positions on the leg and instruments to sub-millimetre accuracy. 
\par
Eichelberger \textit{et al.}  performed an analysis of the accuracy in optical motion capture using trueness and uncertainty of marker distances for human kinematics \cite{eichelberger2016analysis}. The optical volume setup is essential and they showed that the accuracy of the system is highly influenced by the number of cameras and movement conditions \cite{eichelberger2016analysis}. Both these observations are key considerations during robotic surgery and a motivation in this study to customised the RBs and optical volumes. 
\par
\section{Rigid Body Design}
\label{RBDesigns}
The femur anatomical axis follows the femoral head and femur structures, while the femoral mechanical axis (FMA) is the axis that links the hip ball joint centre to the centre of the condyles on the knee. The FMA determines the hip to knee joint motion, even though tracking devices are mounted to the femur.
\par
For automation and to minimise interference in the surgical area, the patient's leg is moved robotically from the heel position as shown in \fig{LegRobot} and as currently performed by surgeons. 
Marker data from rigid bodies mounted on the femur and tibia, together with CT scans of the leg, are used to determine positions relative to the anatomy of the leg. We considered the following criteria for optimal rigid body designs and marker setup:
\begin{enumerate}
  \item Maximise marker visibility during an arthroscopy
  \item Markers from OptiTrack need to fit the RBs
  \item No damage to RBs due to surgery
  \item Fit to existing surgical pins and instruments 
  \item Optimal size, material and shape
  \item The system needs to have a positional accuracy in the sub-millimetre range locally and across joints
  \item Support setup of dynamic (real-time) frames
 \end{enumerate}
\subsection{Femur and Tibia Rigid Bodies}
\begin{figure}[ht]
\centering
    \subfloat[Mounting Plate]{\label{LegRigidBodyplate}{\includegraphics[height=3.25cm]{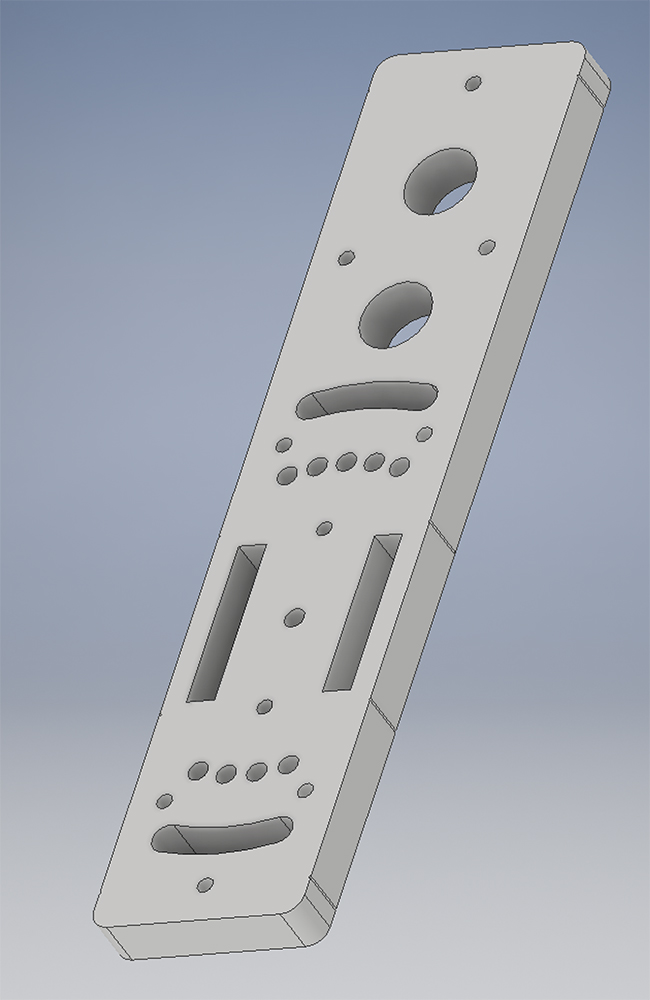}}}
    ~
    \subfloat[Assembled leg rigid body with markers installed on leg and robot mountable foot plate]{\label{femurRB}{\includegraphics[width=0.58\linewidth]{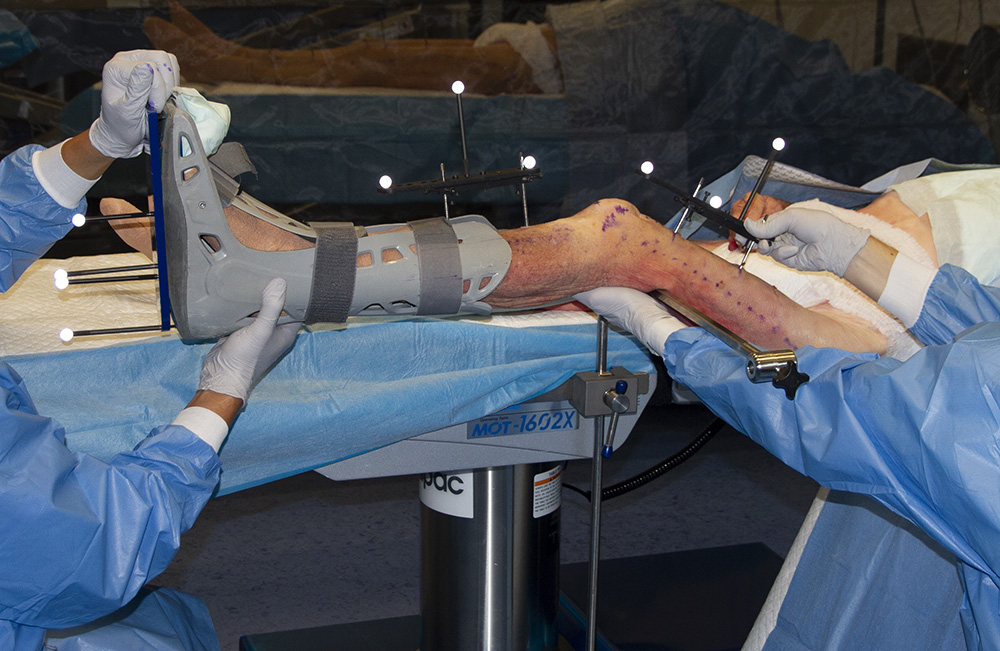}}} \\   
    ~
    \subfloat[Arthroscope Adaptor]{\label{ScopeRB}{\includegraphics[height=2.8cm]{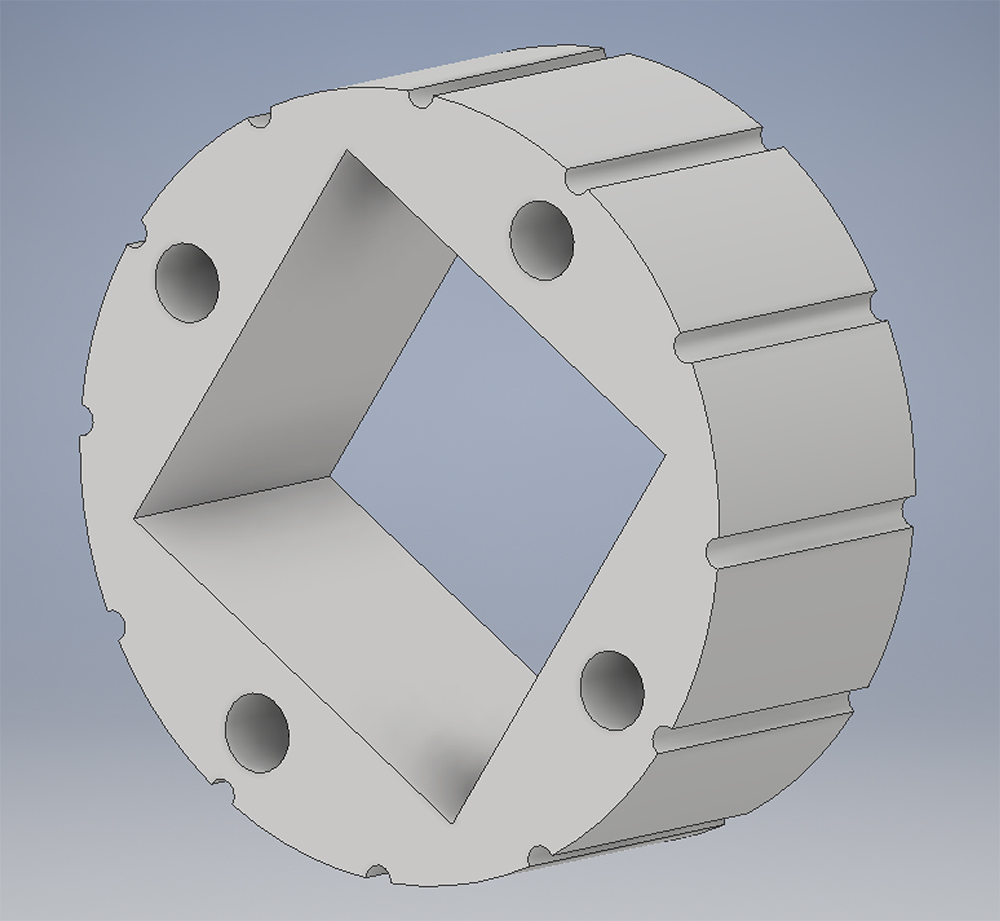}}}
    ~
    \subfloat[Rigid body for arthroscope tracking]{\label{scopeRigidBody}{\includegraphics[width=0.48\linewidth]{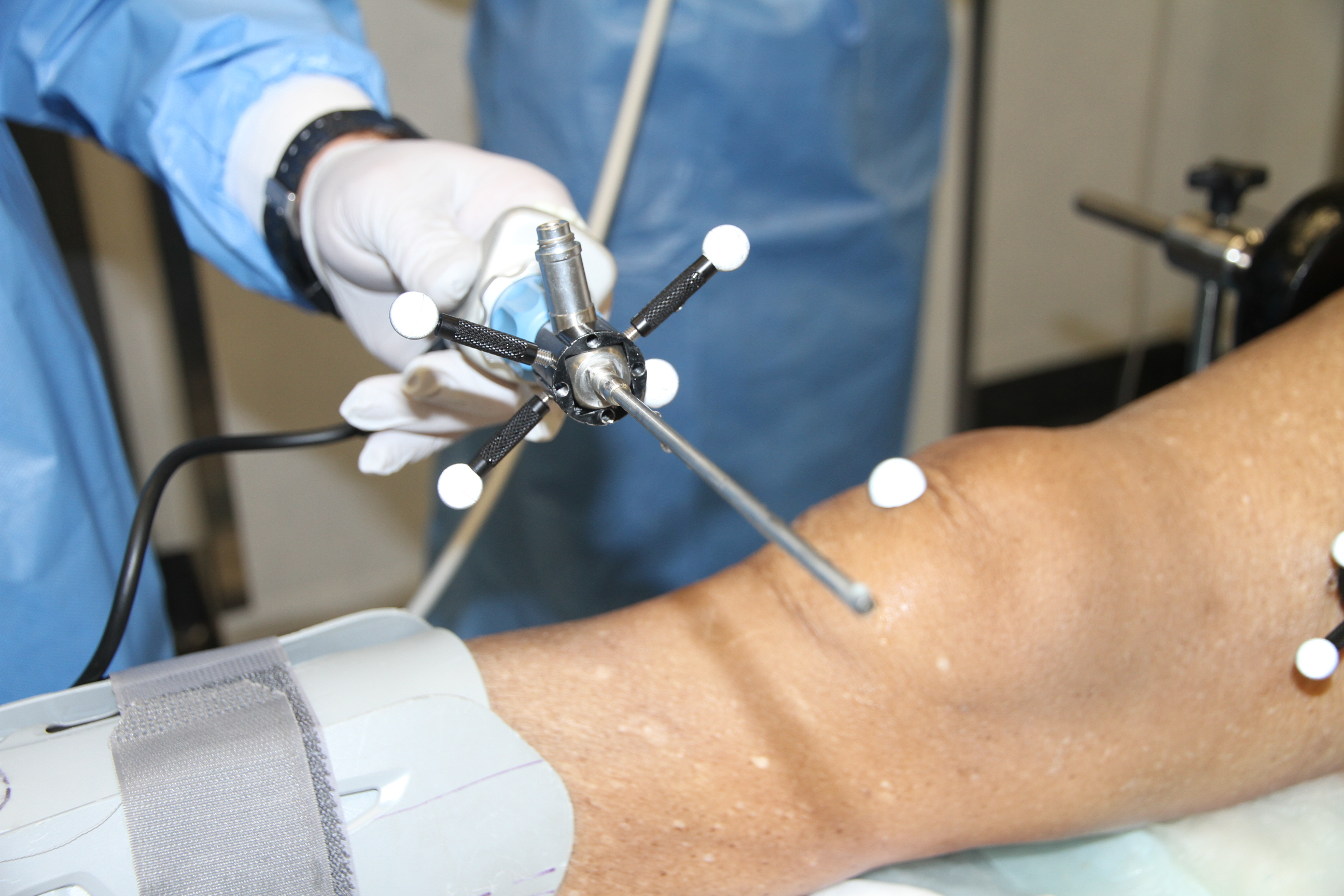}}}
     	\caption{Rigid Bodies on Leg and Arthroscope}
	\label{ScopyBody}      
\end{figure}

From experimental trials on artificial and cadaver legs and joints, various rigid bodies were developed for both the femur and tibia (\fig{ScopyBody}). Trialling a few designs allowed placement of markers in several positions, until the RBs were small enough to fit on the tibia or femur and didn't interfere with surgical manipulation. 
The RB marker plate (\fig{LegRigidBodyplate}) for the femur or tibia have limited (5mm) adjustment when mounted, to allow alignment on the leg. A mounting base is attached to the pins with the marker plate that fits on the base plate, as shown in \fig{femurRB}. The marker plate adjusts relative to the base plate in all directions relative to the leg. Once the markers are installed on the plate, it forms a rigid body that can be tracked in real-time to support analysis of leg motion, as shown in \fig{femurRB}.
Mounting a rigid body with markers on the femur or tibia for this study required the use of existing surgical pins and drilling two of them through into of the bones (\fig{femurRB}) to ensure a solid fixture. As part of this study we added markers on the robot boot (Boot RB that is rigid relative to the tibia), which can be used instead of the tibia RB to track positions in the lower leg or foot. For this study, results are only showed with the RB on the tibia.
\subsection{Rigid Body for Surgical Instruments}
Tracking of surgical camera/instruments is significant for autonomous navigation, monocular measurements or 3D reconstruction of the cavity. The proposed design of the arthroscope marker (\fig{ScopeRB} and \ref{scopeRigidBody}) is based on experimenting with standard OptiTrack RBs during cadaver surgeries and improving on rigidity, size and placement of the markers for continuous tracking during surgery.
The rigid body on the arthroscope has a square, as shown in \fig{ScopeRB} that tightly fits onto the arthroscope. The complete assembly was tested during a cadaver experiment, as shown in \fig{scopeRigidBody}. The markers are positioned such that they don't obstruct the motion of the instrument or interfere with the surgeon.
\section{Optical Volume}
\begin{figure}[ht]
\centering
    \subfloat[Optical volume setup with cameras mounted above (marked in yellow) on a frame around the theatre bed.]{\label{Volume}{\includegraphics[width=0.69\linewidth]{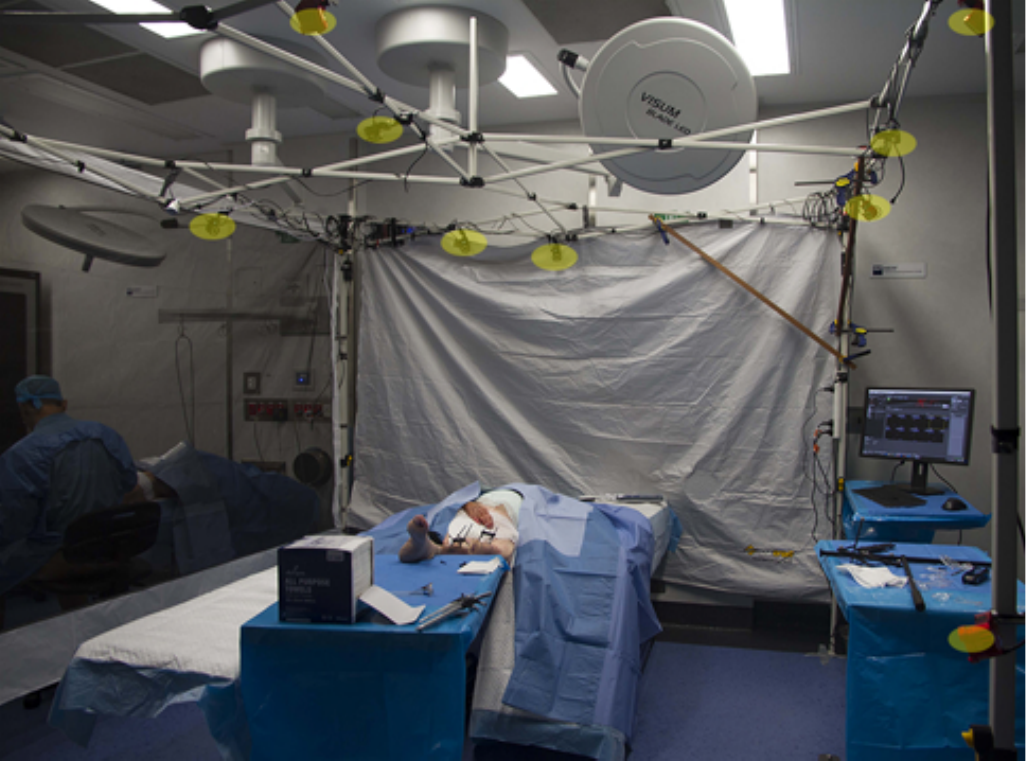}}}
    ~
    \subfloat[At least 3 markers always visible inside volume.]{\label{VisibleMarkers}{\includegraphics[width=0.29\linewidth]{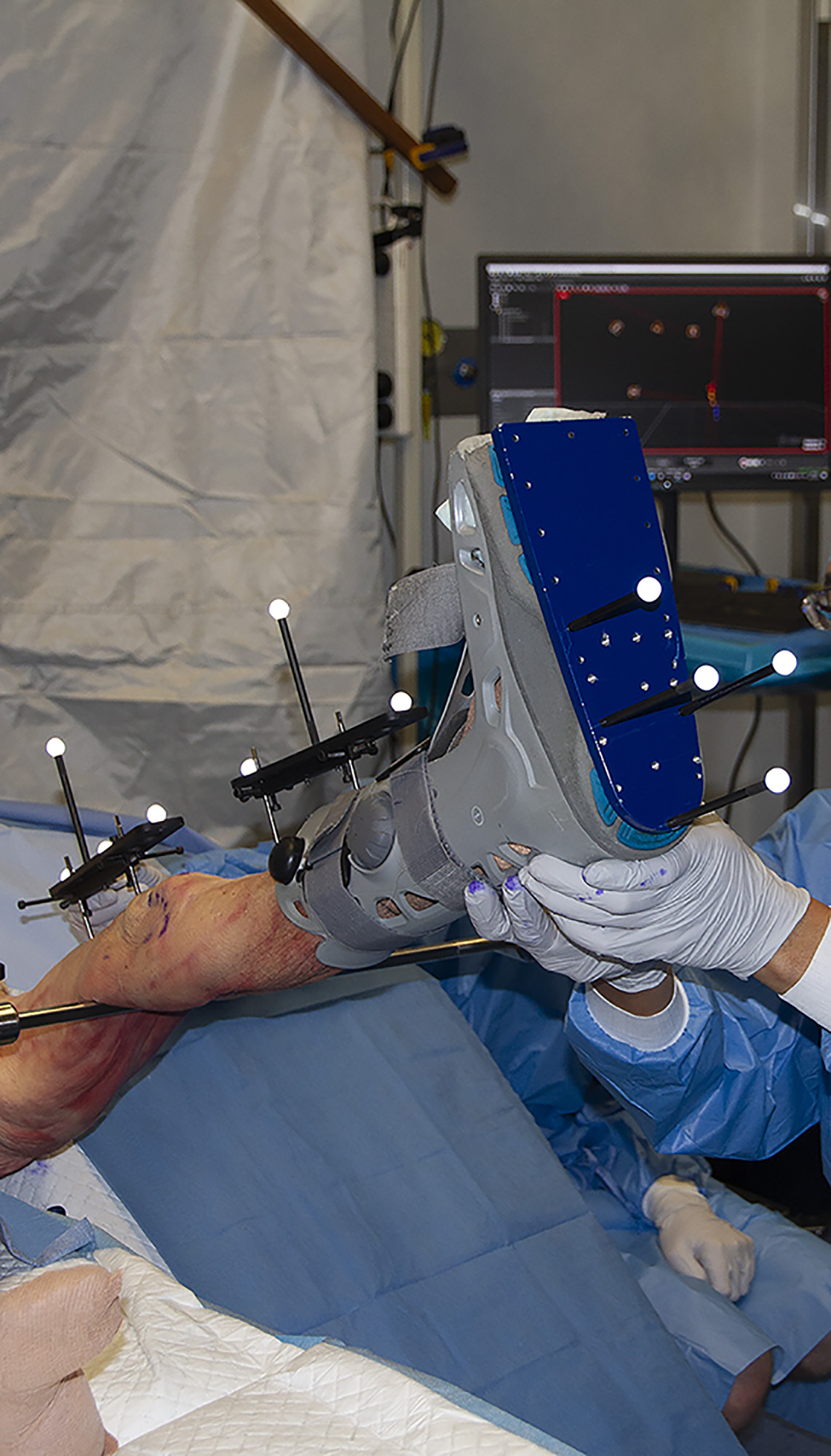}}}\\
         \caption{Optical Volume setup}
    \label{MarkerInVolume}      
\end{figure}

The optical volume setup determines the tracking accuracy. To effectively reconstruct the RB layout (if some markers is occluded) at least three markers (can be different markers over time) need to be visible from three cameras at all times, irrespective of staff and equipment placing. Marker and RB layout can increase visibility, however increasing the number of cameras achieves a higher accuracy for marker location, and more markers can be tracked during surgical manoeuvres. An optimal setup was tested where cameras were setup above the theatre bed and placed on all four corners and the centre of the bed as shown in \fig{Volume}. Together with the unique RB designs, tracking was never lost during any leg motion (\fig{VisibleMarkers}).
\section{Lower leg Motion Analysis}
In order to estimate poses of any chosen point on or inside the leg, it is necessary to setup coordinate frames on key position of a rigid body mounted on the leg. In knowing the position of the optical markers with respect to the OptiTrack global frame (W) and the CT images (\fig{FHCTScan}), it is possible to calculate the local transformation between the RBs and points on the leg. It support the retrieval of the pose of any position on the leg with respect to the global frame (W). In this section calculus behind the described concept is formulated and leg joint angle are calculated to demonstrate its practical use.
\subsection{Marker Coordinate Frames}
\begin{figure}[ht]
\centering
    \subfloat[Markers on Femur Rigid Body RB1. Marker G is closest to the body, H is near the knee.]{\label{RBMarkerSetup}{\includegraphics[width=0.62\linewidth]{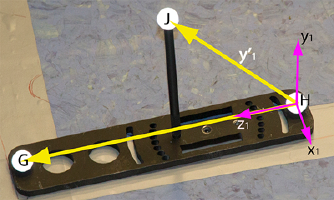}}}
    ~
    \subfloat[CT Scan slice of femoral head, with measurements from the RB1.]{\label{FHCTScan}{\includegraphics[width=0.36\linewidth]{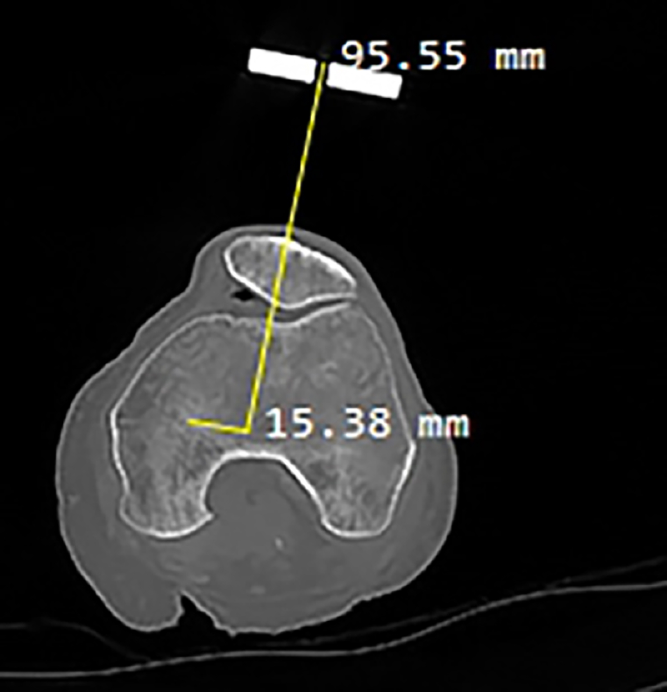}}}\\
         \caption{Optical Volume setup}
    \label{RBCT}      
\end{figure}

Instrument and leg pose analysis requires the setup of frames from marker information measured during optical tracking, using the rigid body designs as detailed in section \ref{RBDesigns}. The axis for the analysis uses a Y-up right-hand coordinate system to align for the optical tracking system configuration, as shown on marker H (\fig{RBMarkerSetup}).
 \begin{figure*}[ht]
	\centering
	\vspace{3 mm}
	\includegraphics[width=0.7\linewidth]{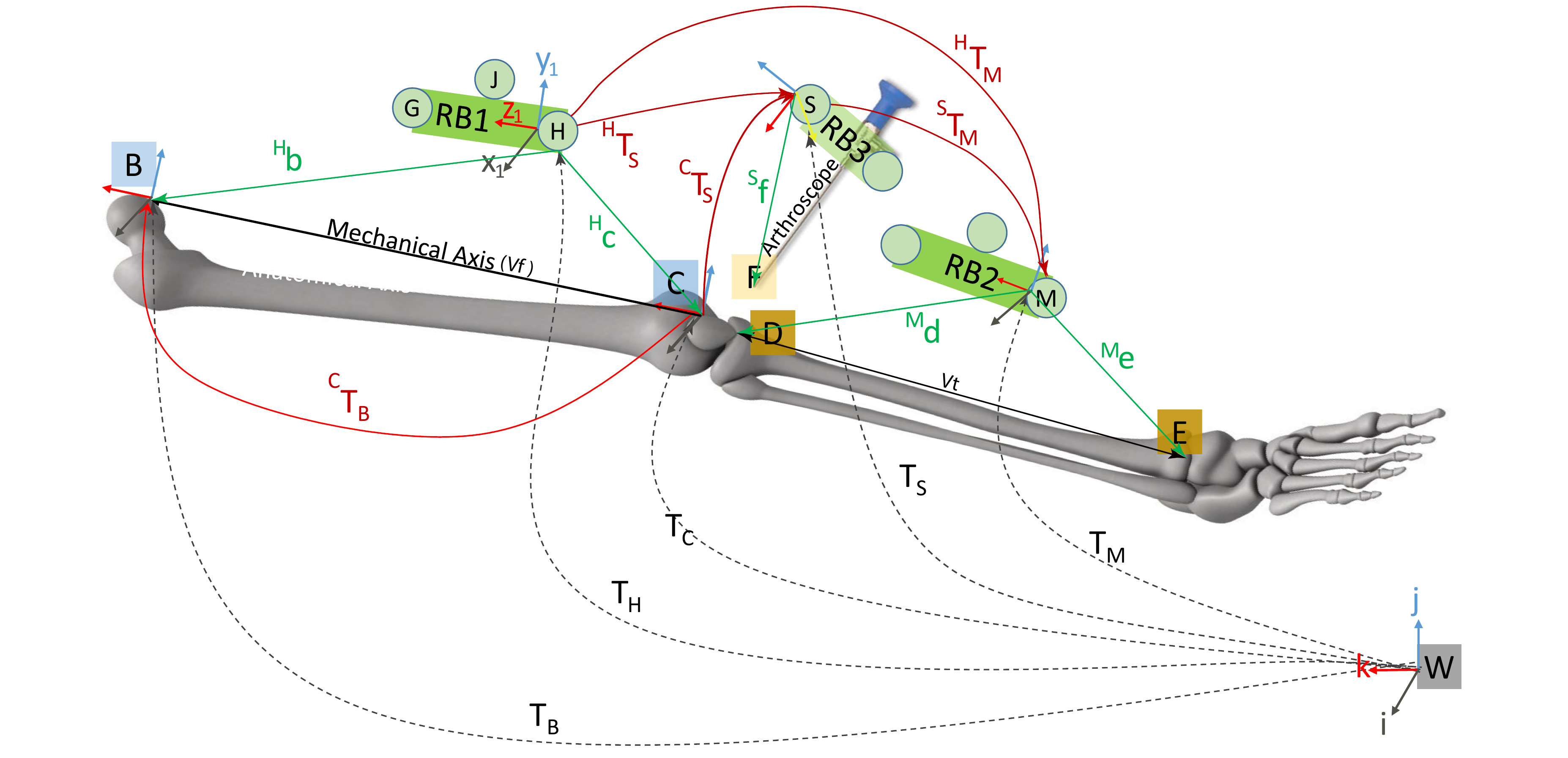}
	\caption{Transformations and vector on the Human leg for an Arthroscopy.}
	\label{Transformations}       
\end{figure*}
The generalised homogeneous transformation matrix (using notations as detailed in \cite{Corke2011}) of the marker G coordinate frame, relative to the origin (or pose of frame H relative to frame W - see \fig{Transformations}) is:
\begin{equation}
\boldsymbol{T}_{H} =
\begin{tikzpicture}[baseline=(m.center)]
  \matrix (m) [
      matrix of math nodes,
      left delimiter={[},
      right delimiter={]},
      row 1/.style={nodes={text height=1ex}}
    ] {
      \textit{x}_i & \textit{y}_i & \textit{z}_i & \textit{$^{w}{\boldsymbol{h}}$}_i\\ 
      \textit{x}_j & \textit{y}_j & \textit{z}_j & \textit{{$^{w}{\boldsymbol{h}}$}}_j\\ 
      \textit{x}_k & \textit{y}_k & \textit{z}_k & \textit{{$^{w}{\boldsymbol{h}}$}}_k\\ 
      0 &0 & 0 & 1\\
    };
    \node [right=1ex of m-2-4] {\quad\quad$\in${$\boldsymbol {SO(3)}$}$\subset\mathbb{R}^{3}$};
  \end{tikzpicture}
  \label{eq:1}
\end{equation}
where \textit{x}, \textit{y} and \textit{z} (first three columns)  are the local frame axes on the rigid body at point H and \textit{i}, \textit{j} and \textit{k} the unit vector axes of the global frame (W). For a frame on marker H (RB1 in \fig{Transformations}), the axes for the transformation matrix (T) can be calculated directly from the rigid body using marker combinations to create vectors between points that align with the rigid body as shown in \fig{RBMarkerSetup}:
\begin{enumerate}
  \item The RB1 z-axis (\textit{$z_i,z_j,z_k$}) is a unity vector from H to G
  \item The frame x-axis (\textit{$x_i,x_j,x_k$}) is:
 $x=y'\times z$ 
  \item The y-axis (\textit{$y_i,y_j,y_k$}) is: 
  $y=z\times x$
  \item The position vector ($\textit{{$^{w}{\boldsymbol{h}}$}}_i,\textit{{$^{w}{\boldsymbol{h}}$}}_j, \textit{{$^{w}{\boldsymbol{h}}$}}_k$) is the marker position relative to W.
 \end{enumerate}
 Using the homogeneous matrix (\ref{eq:1}), we are able to setup frames on any of the markers of any rigid body. For example, the transformation $T_{B}$ defines the pose of a frame on an anatomical point on the femur (B) relative to the world frame (W).
\subsection{Local Transformations}
 A CT scan of the leg is essential to determine the vectors for any point of the leg with respect to one of the marker frames. It is beneficial to perform the CT scan in various positions to enable measurements of different points of the leg as shown in \fig{FHCTScan}, where the measurements was taken for the local translation from RB1 to the centre of the femoral head.
 \fig{MoFrame} shows a CT scan of the femur and the relationship between the mechanical and anatomical axes of rotation of the femur relative to the hip.
Using dynamic frames on the leg, we can determine any positions on the leg or arthroscope at any point in time, and relative to a specific frame. For instance, point C (or vector from W to C) on the leg relative to W is:
\begin{align}
    ^{w}{\boldsymbol{c}} &= {\boldsymbol T}_{H}\quad ^{H}{\boldsymbol{c}} \label{eq:2}
\end{align}
Where $^{H}{\boldsymbol{c}}$ are the local translation (vector) from frame H on RB1 to C on the leg (green lines in \fig{Transformations}).
\subsection{Transformations between legs and Instruments Coordinate Frames}
The transformation between rigid bodies can be determined from the relationship between frames on the RBs or leg. As an example, for the transformation from frame M to frame H:
\begin{align}
    ^{H}T_{M} &= T_{H}^{-1}\, T_{M}\label{eq:4}
\end{align}
Any point on the tibia in frame M can thus be expressed relative to frame H on the femur. One of the key outcomes of this study is that we can express any point relative to any frame, even across joints.
\subsection{Arthroscope Tip Position}
\label{PointsRel2Frame}
To know in real-time the arthroscope tip position in the femur C frame ($^{C}{\boldsymbol{f}}$) we observe from \fig{Transformations} that:
\begin{align}
    ^{C}{\boldsymbol{f}} &=^{G}T_{C}^{-1}\;T_{G}^{-1}\;T_{S}\;^{S}{\boldsymbol{f}}
    \label{eq:6}
\end{align}
\subsection{Motion Analysis}\label{Motion}
A typical surgery is performed with the patient lying on their back and for this study, we choose y-up and z aligned along the body from toe to head.
Using the transformations described above, we define vectors between points on the bones, from which knee and hip joint rotations and translations are analysed.
\subsubsection{Knee Angles}
\begin{figure}[ht]
	\centering
	\includegraphics[width=\linewidth]{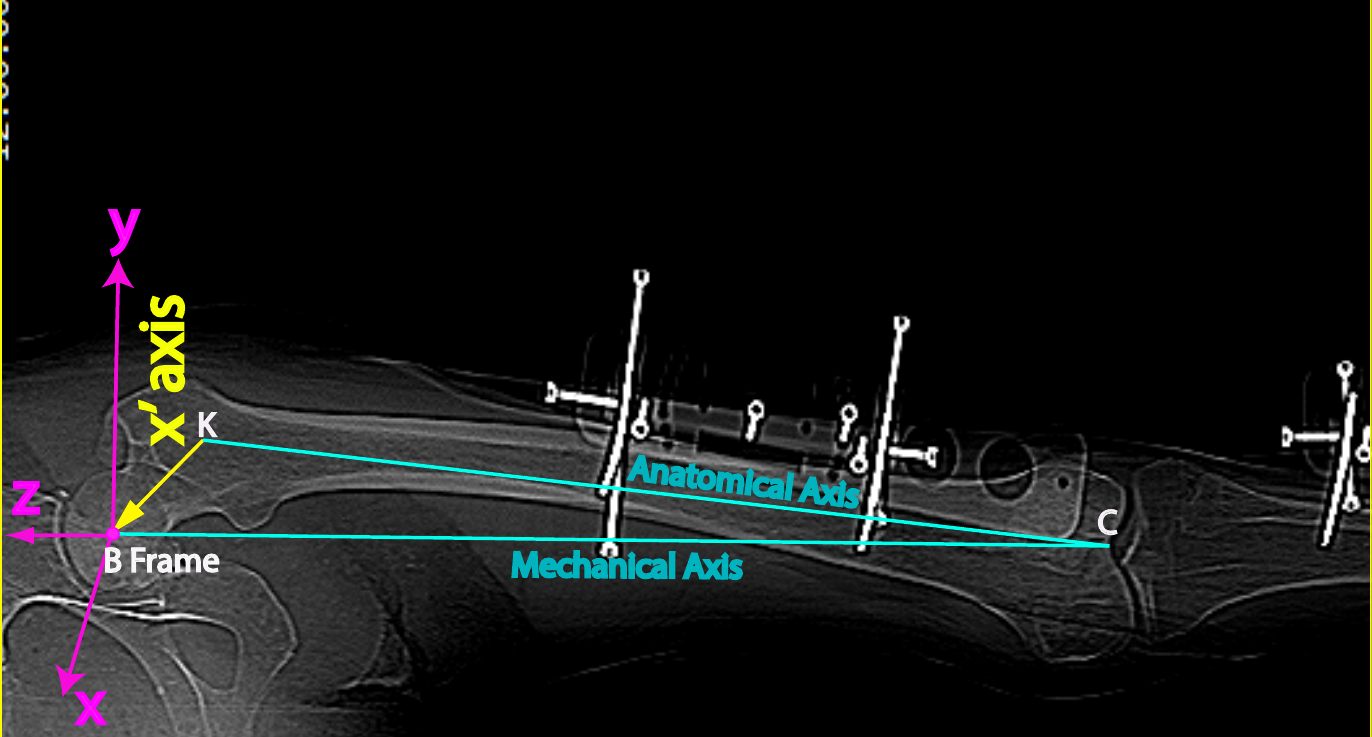}
	\caption{From CT Scan the B Frame (pink) setup on femoral mechanical axis, using ball joint to femur (point K) to determine the frame's x and y axis.}
	\label{MoFrame}       
\end{figure}

The tibia vector is defined from the centre of the condyles to the ankle centre. However, the motion of the tibia is relative to the femur (rotations and translations) and needs to be measured relative to a frame on the femoral condyle centre. The rotational matrix on the condyle can be setup using measurements using the CT scan data as shown in \fig{MoFrame}. 
As illustrated we determine the centre of the ball joint (B) and the connection centre (K) of the femoral anatomical and femoral head axes. This vector forms the x' axis and we obtain:
\begin{align*}
{\boldsymbol{z}}&=^{w}{\boldsymbol{c}-^{w}}{\boldsymbol{b}}   \quad\text{frame z axis} (mechanical)\\
{\boldsymbol{x}'}&=^{w}{\boldsymbol{b}-^{w}}{\boldsymbol{k}}   \quad\text{x' axis}\\
\boldsymbol{y}&=\boldsymbol{z}\times\boldsymbol{x'}   \quad\text{y axis}\\
\boldsymbol{x}&=\boldsymbol{y}\times\boldsymbol{z}   \quad\text{x axis}    
\end{align*}
We define the zx' plane by points B, K and C in \fig{MoFrame} with y perpendicular to this plane. The rotational frame ($^{W}\boldsymbol{R}_{B}$) on the FMA is the combination of the x, y and z vectors on point B. For rotations or translations of the tibia relative to the femur, the transformation frame in point C, on the femoral mechanical axis is:
\begin{equation}
\boldsymbol{T}_{C} =
\begin{tikzpicture}[baseline=(m.center)]
  \matrix (m) [
      matrix of math nodes,
      left delimiter={[},
      right delimiter={]},
      row 1/.style={nodes={text height=1ex}}
    ] {
      \boldsymbol{x} & \boldsymbol{y} & \boldsymbol{z} & ^{w}{\boldsymbol{c}_B}\\ 
      0 &0 & 0 &1\\
    };
  \end{tikzpicture} 
  =
  \begin{tikzpicture}[baseline=(m.center)]
  \matrix (m) [
      matrix of math nodes,
      left delimiter={[},
      right delimiter={]},
      row 1/.style={nodes={text height=1ex}}
    ] {
      \boldsymbol{R}_{B} & ^{w}{\boldsymbol{c}_B} \\ 
      0 & 1\\
    };
  \end{tikzpicture}
   \label{eq:5}
  \end{equation}
Where $^{w}{\boldsymbol{c}_B}$ is point C in W via frame B. The vector from the centre of frame $\boldsymbol{T}_{C}$ to point E describes the motion of the tibial mechanical axis, which is:
\begin{align}
{\boldsymbol{v}_t}&= ^{C}{\boldsymbol{e}_M}  \quad\text{(Tibia Vector)}  \label{eq:8a}
\end{align}
In kinematics the angles of the hip and knee joints are extensively used, and is essential for future robotic applications. For this study we will use the rigid body system to calculate the joint angles and use synchronised video from the OptiTrack system to visually compare the results. Using vector analysis the knee varus ($\beta$) and flexion ($\alpha$) angles can be calculated:
\begin{align*}
     \vect{v_{t_x}} &= \vectorproj[x_n]{v_t} = \frac{\vect{v_t} \cdot \vect{x_n}}{\norm{\vect{x_n}}^2} \vect{x_n}\\
     \vect{v_{t_{yz}}} &= \vect{V_{t}}-\vect{V_{t_x}}
\end{align*}
$\vect{v_{t_x}}$ is the projected $\vect{v_t}$ vector on the unity vector ($\vect{x_n}$) of the femur C frame's x-axis and $\vect{v_{t_{yz}}}$ the $\vect{v_t}$ vector in the yz-plane. Using these vectors we can calculate the dot and cross product between $\vect{v_{t_{yz}}}$ and $\vect{v_t}$, with the knee varus angle:
\begin{align}
     \beta &= \atantwo ( \norm{\vect{v_{t_{yz}}}\times\vect{v_{t}}},\vect{v_{t_{yz}}} \bigcdot \vect{v_{t}})\label{eq:9a}
 \end{align}

Projecting $\vect{v_t}$ to the xz plane, the knee flexion angle is:
\begin{align}
     \alpha &= \atantwo ( \norm{\vect{v_{t_{xz}}}\times\vect{v_{t}}},\vect{v_{t_{xz}}} \bigcdot \vect{v_{t}})\label{eq:10a}
 \end{align}
 Using a rotational matrix is an alternative option of calculating the knee angles between vectors $v_f$ and $v_t$. The rotational matrix between the femur and tibia is:
\begin{equation*}
^{v_f}\boldsymbol{R}_{v_t} = 2 \frac{(t_{r} t_{r}^{-1})}{(t_{r}^{-1} t_{r})-I}\quad\text{with: } t_{r} = \vect{v_{f}}+\vect{v_{t}}
\end{equation*}
and using the matrix, the knee IE angle $\gamma$:
\begin{align}
     \gamma &= \atantwo ( -^{V_f}\boldsymbol{R}_{v_t}(1,2),^{V_f}\boldsymbol{R}_{v_t}(1,1)) \label{eq:10b}
 \end{align}
\subsubsection{Knee Translations}
During minimally invasive surgery, the knee gap size between the femur and tibia is required for accessing inner knee areas with surgical instruments. Translations in the joints can be measured by setting up vectors at the condyle joint points C and D, that is using point D in frame C (see Section \ref{PointsRel2Frame}). $^{C}{\boldsymbol{d}}$ will provide the x (medial/lateral),y (posterior/anterior) and z (knee gap) translation of the knee joint as a result of rotation and translation during motion.
\subsubsection{Hip Angles}
\label{HipA}
The femur mechanical axis is defined as the link from the hip joint centre to the centre of the condyles on the knee as shown in \fig{MoFrame} and \ref{CTScanAnalysisTop2Bottom}(a). The femur vector that describes the hip rotations relative to the world frame is:
\begin{align}
{\boldsymbol{v}_f}&=^{B}{\boldsymbol{c}}-^{B}{\boldsymbol{b}}   \quad\text{Femur Vector}  \label{eq:11}
\end{align}
Angles and translations are measured relative to the sagittal (flexion), coronal (varus) and transverse (knee gap) planes. Using vectors, the hip varus ($\psi)$ and flexion($\theta$) angles are:
\begin{align}
     \psi &= \atantwo ( \norm{\vect{v_{f_{yz}}}\times\vect{v_{f}}},\vect{v_{f_{yz}}} \bigcdot \vect{v_{f}})\label{eq:12}
 \end{align}
\begin{align}
     \theta &= \atantwo ( \norm{\vect{v_{f_{xz}}}\times\vect{v_{f}}},\vect{v_{f_{xz}}} \bigcdot \vect{v_{f}})\label{eq:36}
 \end{align}
For the hip roll angle we can project $v_{f}$ to the yx plane and calculate the angle between the plane and $v_{f_{yx}}$, however we can also use rotational matrices. Using $^{W}\boldsymbol{R}_{C}$ (\ref{eq:5}) we get the hip roll angle as:
\begin{align}
     \psi &= \atantwo ( -^{W}\boldsymbol{R}_{C}(1,2),^{W}\boldsymbol{R}_{C}(1,1)) \label{eq:14}
 \end{align}
\section{Experimental Validation}
\subsection{Experimental Setup}
\begin{figure}[ht]
\centering
    \subfloat[Femur Anatomical and mechanical axes]{\label{AnatomicalAxis2}{\includegraphics[height=7cm]{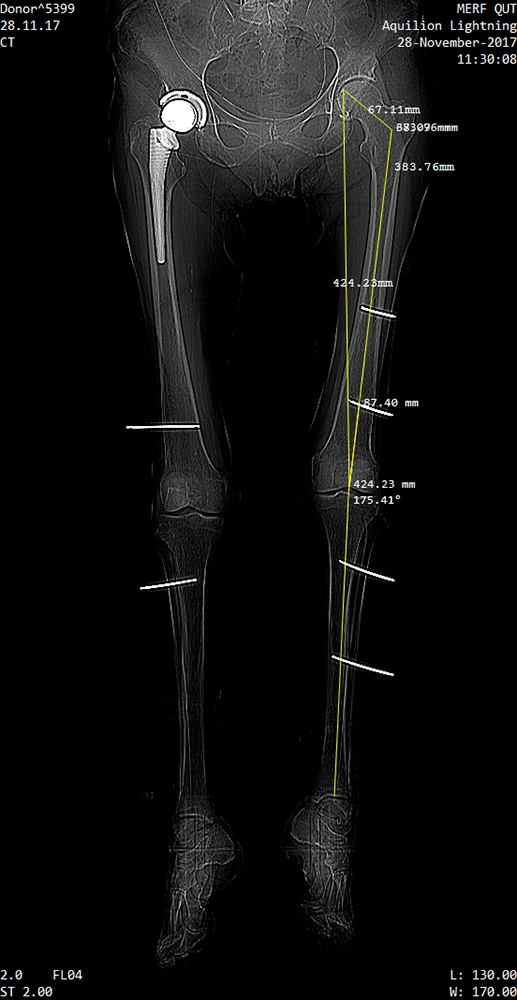}}}
    ~
    \subfloat[Mounted RBs and markers.]{\label{OptiTrackMarkers}{\includegraphics[height=7cm]{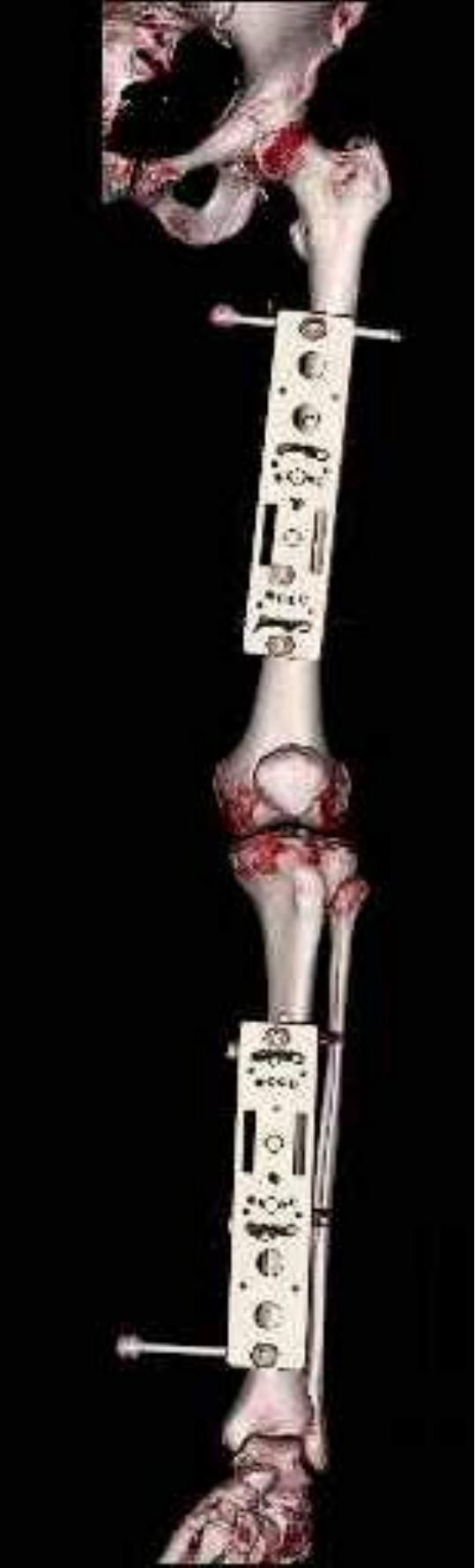}}}\\
  \caption
    {CT Scan of a cadaver leg.
      \label{CTScanAnalysisTop2Bottom}%
    }%
\end{figure}

    A leg manipulator robot (\fig{LegRobot}) was developed (patent No. 2019900476) with a foot interface (\fig{femurRB}) to the leg. Ethical approvals were gained for three Cadaver experiments as detailed in Table \ref{CadaverExperiments}. Firstly the robustness and accuracy of existing rigid bodies from OptiTrack were tested. A ten camera system were installed on a 3mx3mx2.5m structure (\fig{MerfSurgery}), and calibrated with an accuracy wand. The second experiment tested the designed RBs and CT scan measurements as shown in \fig{CTScanAnalysisTop2Bottom}. For the final experiment a refined process, CT scans and RBs designs were used. 

\begin{table}[htbp]
  \caption{Cadaver experiments to test standard rigid bodies from OptiTrack, the newly designed rigid bodies as well as the leg motion.}
  \footnotesize
    \begin{tabular}{cp{9.5em}p{4.04em}c}
    \toprule
    \multicolumn{1}{P{4.675em}}{\textbf{Experiment}} & \textbf{Cadaver} & \textbf{Sex} & \multicolumn{1}{P{3.44em}}{\textbf{Age}} \\
    \midrule
    \midrule
    OptiTrack Std RB   &Left and Right Knees & Male  & 80-90 \\
    \midrule
    Designed RBs    &Left and Right Knees & Male  & 60-70 \\
    \midrule
    Kinematic Tests     & Left and Right Knees & Female & 50-60 \\
    \bottomrule
    \bottomrule
    \end{tabular}%
  \label{CadaverExperiments}%
\end{table}%
A 4mm Stryker arthroscope and an OptiTrack System were used during experiments. The designed RBs were mounted on the cadaver femur, tibia, arthroscope and robot boot. Markers were mounted in specific patterns for real-time visibility and frame setup.
\subsection{Experimental Results}
OptiTrack results show that there is continues visibility of the markers during a full leg motion experiment of 4 minutes. Enough markers were tracked on each RB for the OptiTrack system to fully recover the position of each marker. 
\par
Table \ref{PointDviaB} shows point E relative to the world frame ($^{w}{\boldsymbol{e}_M}$) via local translation from M to point E and then a translation via frame C and M to point E ($^{w}{\boldsymbol{e}_BM}$).
\begin{table}[ht]
  \centering
	\caption{Ankle-E point 1) Directly from the world frame and 2) via frames C to M to E.}
   \footnotesize
    \begin{tabular}{c|ccc|ccc}
    \toprule
    \multicolumn{1}{p{3em}}{\textbf{\hspace{0mm}Time}} &
    \multicolumn{3}{p{7.8em}}{\textbf{\hspace{10mm}$^{w}{\boldsymbol{e}}_c$}} &
    \multicolumn{3}{p{9.03em}}{\textbf{\hspace{1mm}$^{w}{\boldsymbol{e}}_{cme}$}}\\
    \multicolumn{1}{p{2em}}{\textbf{[Sec]}} &
    \multicolumn{1}{p{2em}}{\textbf{Ex}} &
    \multicolumn{1}{p{2em}}{\textbf{Ey}} &
    \multicolumn{1}{p{2em}}{\textbf{Ez}} &
    \multicolumn{1}{p{2em}}{\textbf{Ex}} &
    \multicolumn{1}{p{2em}}{\textbf{Ey}} &
    \multicolumn{1}{p{2em}}{\textbf{Ez}}\\
    \midrule
    \midrule
   00:11.033  &1221.7& 910.22 & 827.47 &1221.6& 909.495 & 827.47 \\
   01:48.492  &933.06& 859.26 &1088.6 &933.06& 859.26 & 1088.6 \\
   02:42.500  &1354.5& 1135.4 & 848.75 &1354.5& 1135.4 & 848.75 \\
   03:40.525 &1323& 1188.6 &1256.8 &1323& 1188.6 & 1256.8 \\
   04:37:517 &1260.3&1064.4 & 835.7 &1260.3 &1064.4 & 835.7 \\
    \midrule
    \bottomrule
    \end{tabular}%
  \label{PointDviaB}%
\end{table}
\begin{table}[ht]
  \centering
	\caption{Local Translation Error length with $^{w}{\boldsymbol{e}}_D$ via frame D, compared to $^{w}{\boldsymbol{e}}_C$, which is via frame C.} 
	   \footnotesize
    \begin{tabular}{c|ccc|c}
    \toprule
    \multicolumn{1}{p{3em}}{\textbf{\hspace{1mm}Time}} &
    \multicolumn{3}{p{10.8em}}{\textbf{\hspace{10mm}$^{w}{\boldsymbol{e}_D}$}} &
    \multicolumn{1}{p{7em}}{\textbf{\hspace{4mm}Error}}\\
    \multicolumn{1}{p{2em}}{\textbf{[Sec]}} &
    \multicolumn{1}{p{2em}}{\textbf{Ex}} &
    \multicolumn{1}{p{2em}}{\textbf{Ey}} &
    \multicolumn{1}{p{2em}}{\textbf{Ez}} &
    \multicolumn{1}{p{7em}}{\textbf{\hspace{2mm}$^{w}{\boldsymbol{e}}_C-^{w}{\boldsymbol{e}_D}$}}\\
    \midrule
    \midrule
   00:11.033  &1221.6& 909.495 & 827.47 & 0.7804\\
   01:48.492 &933.06& 859.26 & 1088.6 &  0.7805\\
   02:42.500 &1354.5& 1135.4 & 848.75 & 0.7935 \\
   03:40.525 &1323& 1188.6 & 1256.8 & 0.7995 \\
   04:37:517 &1260.3 &1064.4 & 835.7 & 0.7807 \\
    \midrule
    \bottomrule
    \end{tabular}%
  \label{PointDError}%
\end{table}
 \begin{figure}[!ht]
\centering
\vspace{3 mm}
    \subfloat[Hip Flexion]{\label{HipFlex}{\includegraphics[width=0.49\linewidth]{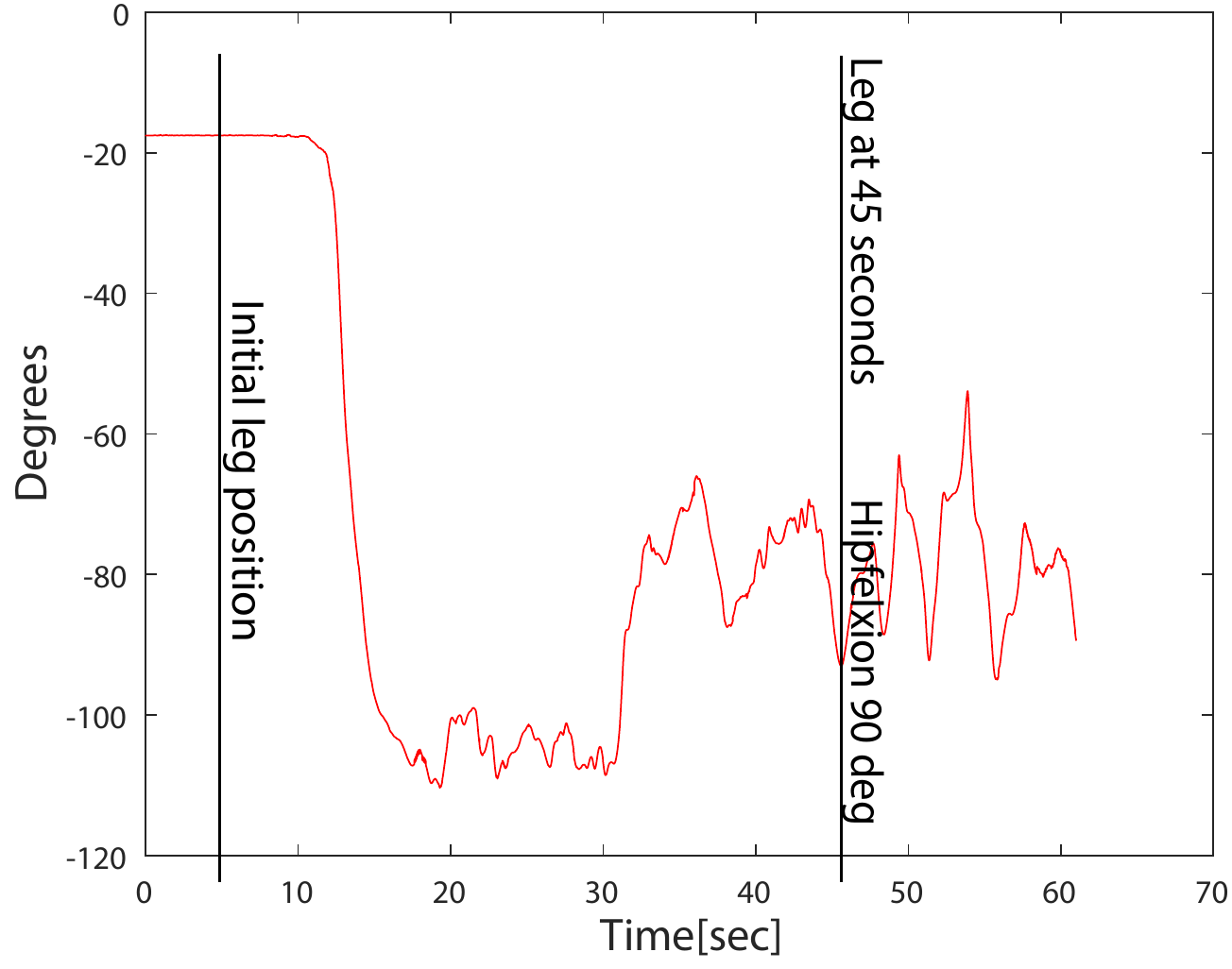}}}
    ~
    \subfloat[Knee Flexion]{\label{KneeFlex}{\includegraphics[width=0.49\linewidth]{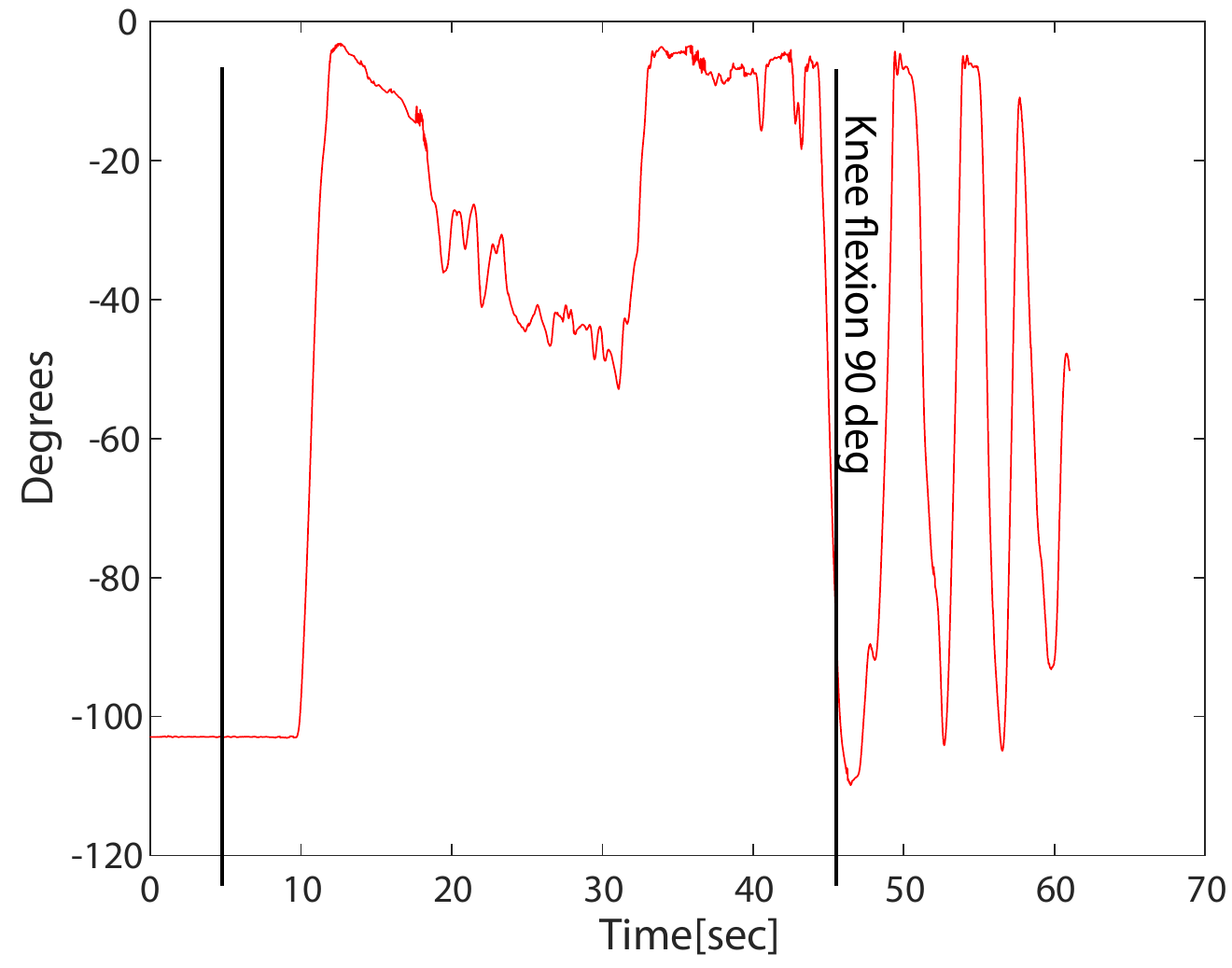}}}\\
    ~
    \subfloat[Hip Varus]{\label{HipVarus}{\includegraphics[width=0.47\linewidth]{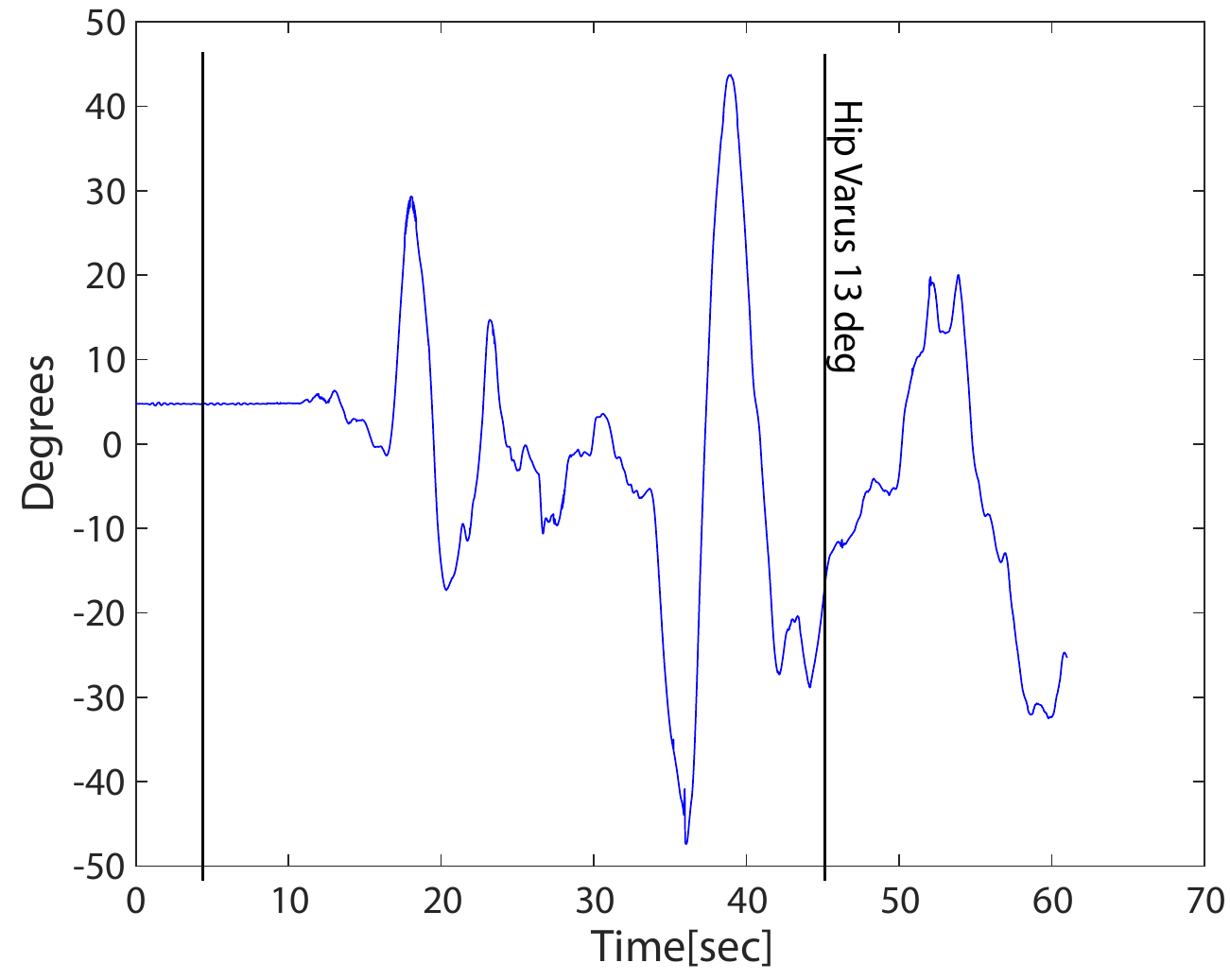}}}
    ~
    \subfloat[Knee Varus]{\label{KneeVarus}{\includegraphics[width=0.47\linewidth]{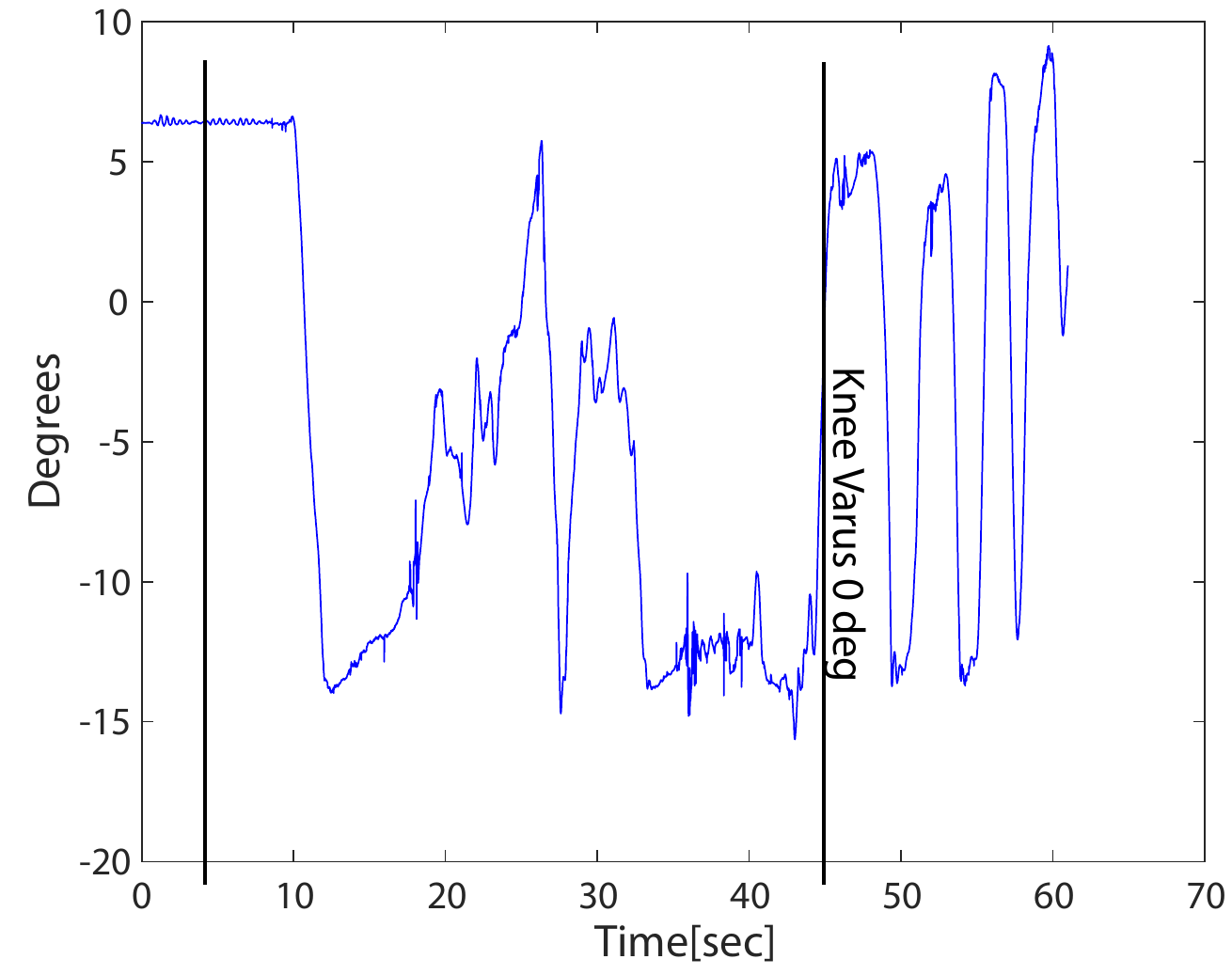}}}\\
        ~
    \subfloat[The leg at initial rest position as shown (at time 5 sec) with the fist line on each graph.]{\label{At5sec}{\includegraphics[width=0.49\linewidth]{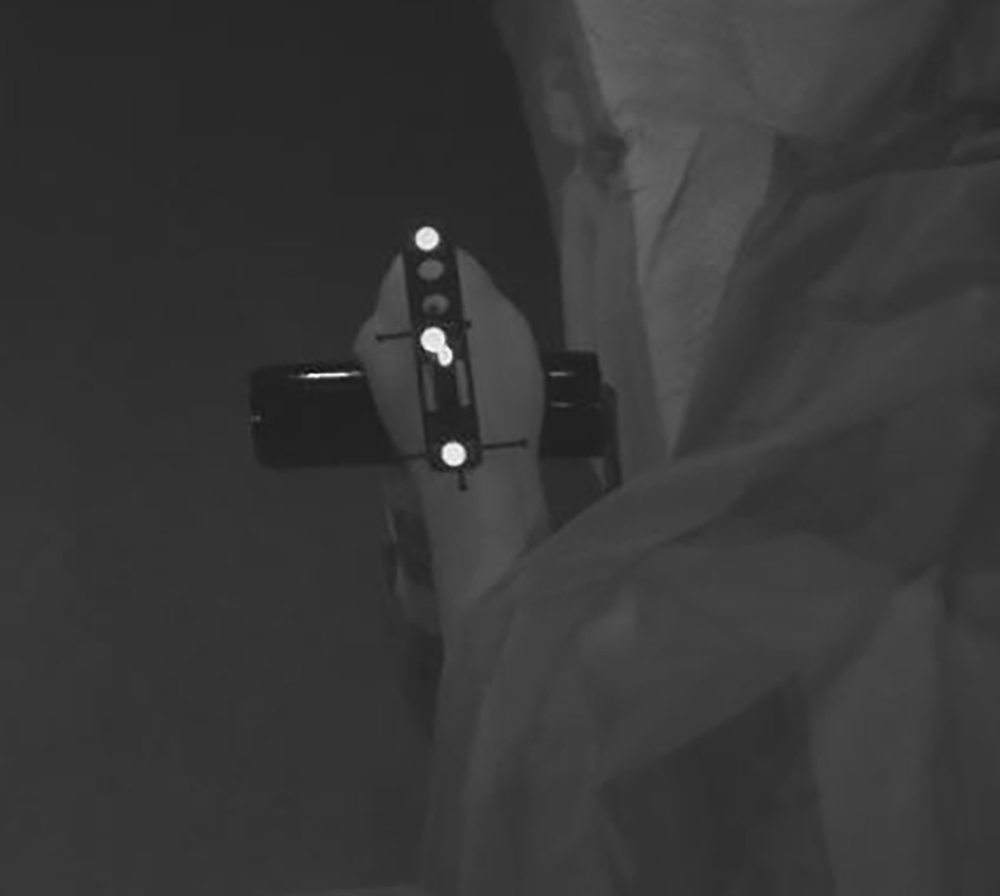}}}
    ~
    \subfloat[The leg at 45 seconds as shown with the second line on each graph.]{\label{At45sec}{\includegraphics[width=0.43\linewidth]{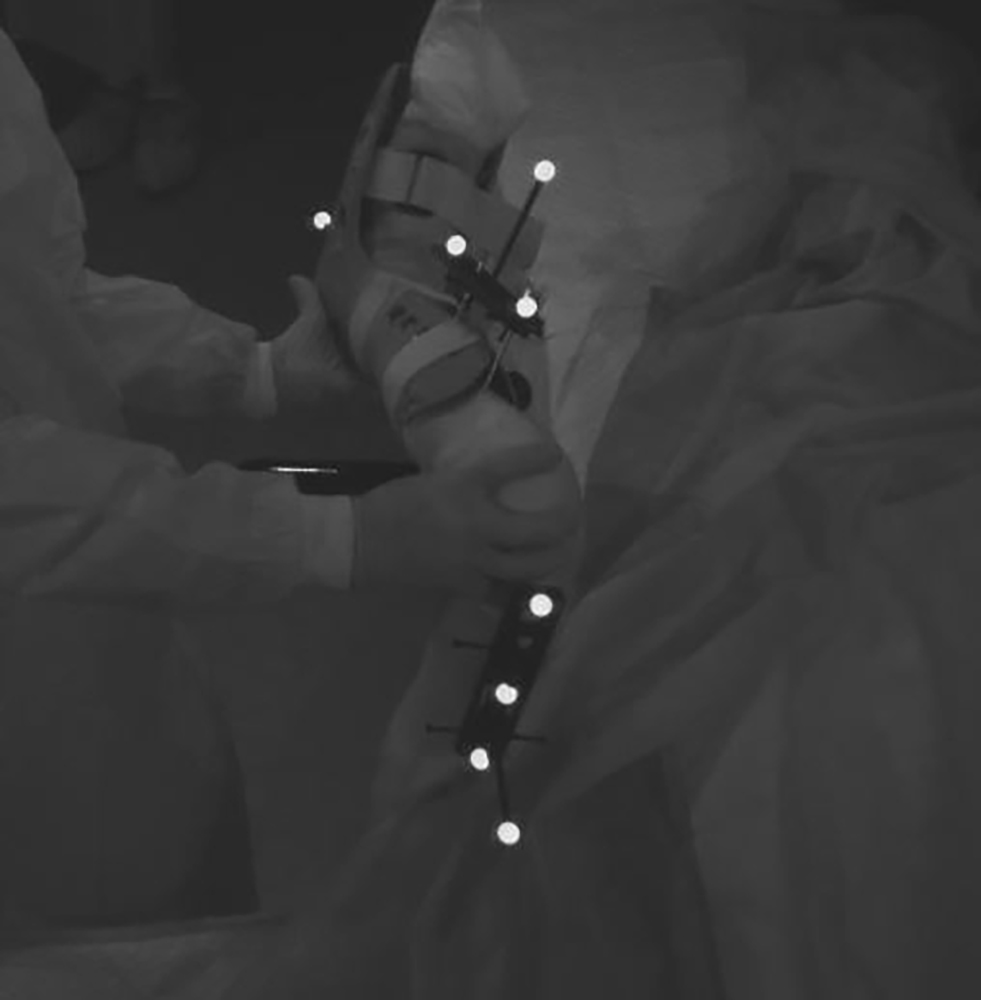}}}\\
\caption{Flexion and Varus Leg Angles - First sixty seconds of a five minute cadaver experiment. The leg was tracked using the designed optical rigid bodies. The positions from the OptiTrack reference video shown in \ref{At5sec} and \ref{At45sec} are at times as indicated on the graphs.}
\label{LegAngles}    
\end{figure} 
Leg angles as shown in \fig{LegAngles} were calculated from the measured marker positions as detailed in Section\ref{Motion} during cadaver experiments. The leg was moved through a range of angles, manually and with the leg manipulator robot.
\section{Discussion}
Providing autonomy for leg positioning and surgical instrument navigation in robotic-assisted orthopaedic surgery requires accurate spatial information. Prior to cadaver experiments, CT scans of the leg were taken and then using the OptiTrack system, marker data was recorded by moving the legs through all possible ranges for leg surgeries. The standard OptiTrack rigid bodies were initially tested and failed physically within a few minutes during the first cadaver arthroscopy. Markers were easily obstructed due to surgeon, staff, patient and instruments motion and manually setting up of frames on specific markers difficult. Rigid body pose data provided by the OptiTrack system is not accurate for multiple leg and instrument tracking, as it relies on manually initialising the rigid bodies with the world frame setup during calibration. 
\par
For a knee arthroscopy, millimetre accuracy is required for measurement of the internal joint parameter such as the size of the knee joint gap needed for the 4mm arthroscope to pass through it. Surgeons regularly overestimate the gap resulting in unintended damage. From testing, the OptiTrack accuracy was found to be 0.03mm when measured over 33000 samples in dynamic surgical conditions and similar to that reported by Maletsky \cite{Maletsky2007}. The positional accuracy of the OptiTrack and the custom rigid bodies for each part of the leg and instruments, ensure real-time data reliability during the surgery. It supports an accurate setup of frames to track points on the leg or instruments. The accuracy of local points on the leg is dependent on the accuracy of the combination of the OptiTrack and CT scan measurements. With CT scan measurement accuracy of 0.3mm \cite{kim2012accuracy}, the accuracy of a point in the leg is largely dependent on that. As shown in Table \ref{PointDError}, the overall accuracy crossing two local measurements is on average 0.75mm, aligning with the CT scan accuracy, which is small relative to sizes in the knee joint and negligible when calculating parameters such as joint angles.
\par
The volume setup of the optical system is critical for visibility. At least three markers on a RB needs to be visible to know all marker positions. It was found that for an arthroscopy ten cameras placed above and at the sides of the volume, ensured continuous optical marker tracking, irrespective of surgeon or instrument motion. For automated leg manipulation or future robotic surgery, movement around the patient is reduced, and fewer cameras and a smaller volume will be required.
\par
The optical tracking accuracy of markers on the leg using the mathematical model is shown in table \ref{PointDviaB}, where the ankle centre point (E) is tracked across RBs, showing consistent positional information for the ankle. The combination of CT and optical tracking shows that during surgery, it is possible to accurately and in real-time translate to points across joints and express points in a joint relative to any frame. For other areas of the body or for different surgeries, it will be necessary to customise the RBs. However, the measurement and mathematical approach remain the same.
\par
Key parameters for robotic leg manipulation include the rotations and translations of each joint, which is calculated from the combination of CT, optical tracking and the mathematical model. It forms an integrated system during surgery for real-time anatomical measurements. Angles for each joint were calculated from the cadaver data and are shown in \fig{HipFlex} to \ref{KneeVarus}. \fig{At5sec} and \ref{At45sec} show snapshots from video analysis at time 5 and 45 seconds, which is marked with black vertical lines on each of the angle graphs. For clarity, only the first 60 seconds are shown. The accuracy of the vector's positional data (0.3mm), ensures that the calculated angles are accurate. 
\par
 For knee surgery, the dynamic space in the knee joint and an arthroscope diameter of 4mm,  make the sub-millimetre accuracy in this study suitable for robotic leg manipulation and instrument guidance. Other applications include modelling of the joint surfaces and structures and alignment of femur and tibia axes.
\section{Conclusion}
Optical marker tracking, customised rigid body designs, CT measurements of anatomical points and kinematic analysis of joint parameters presented in this study, form the first integrated system for leg manipulation support in robotic-assisted orthopedic surgery. During three cadaver experiments, the leg was moved through surgical positions to provide the full motion ranges for the hip and knee joints. The system was verified by translating to known markers across joints. The rotations of the hip and knee joints are calculated, with an accuracy relative to the accuracy of the positional data of the mechanical vectors, which is 0.3mm. The proposed framework has potential to support future robotic-assisted leg surgery by providing real-time data to measure joint, leg and instrument parameters. To reduce patient trauma the foot rigid body can be used to analyse point in the tibia. 
This study creates the basis for future work to develop kinematic models of the human leg, using robotic models such as the DH parameters or to 3D model the knee joint surfaces.
\section{Acknowledgements}
 Cadaveric experiments were approved  by  the  Australian  National  Health  and  Medical Research Council (NHMRC) - Committee no. EC00171, Approval no. 1400000856. The authors would like to thank Fumio Sasazawa and Andres Marmol-Velez for their assistance during cadaver tests. 
\bibliographystyle{IEEEtran} 
\bibliography{RigidBodies}

\end{document}